\documentclass[10pt,twocolumn,letterpaper]{article}

\usepackage{cvpr}              

\usepackage[dvipsnames]{xcolor}

\newcommand{\myparagraph}[1]{\vspace{1.5pt}\noindent{\bf{#1}}}

\newcommand{\B}[0]{\mathbf{B}}
\newcommand{\bH}[0]{\mathbf{H}}
\newcommand{\F}[0]{\mathbf{F}}
\newcommand{\I}[0]{\mathbf{I}}
\newcommand{\V}[0]{\mathbf{V}}

\usepackage{soul}
\usepackage[accsupp]{axessibility} 

\newcommand{\mapping}[1]{{\color{darkgray} \fontfamily{cmss}\selectfont#1}}
\newcommand{\mgrey}[1]{{\color{gray} \fontfamily{cmss}\selectfont#1}}

\usepackage[labelfont=bf]{caption}
\usepackage{arydshln}
\usepackage{stmaryrd}

\definecolor{cvprblue}{rgb}{0.21,0.49,0.74}
\usepackage[pagebackref,breaklinks,colorlinks,citecolor=cvprblue]{hyperref}

\def\paperID{2243} 
\def\confName{CVPR}
\def\confYear{2024}

\title{Self-Supervised Multi-Object Tracking with Path Consistency}

\author{Zijia Lu\thanks{Currently at Northeastern University. Work conducted during an internship with AWS AI Labs.}, \hspace{3pt}  Bing Shuai\thanks{Corresponding author.} , \hspace{3pt} Yanbei Chen, \hspace{3pt} Zhenlin Xu, \hspace{3pt} Davide Modolo\\ 
AWS AI Labs \\
{\tt\small lu.zij@northeastern.edu, $\;$ \{bshuai,yanbec,xzhenlin,dmodolo\}@amazon.com}}

\begin{document}


\maketitle
\vspace{-5mm}

\begin{abstract}

In this paper, we propose a novel concept of path consistency to learn robust object matching without using manual object identity supervision.
Our key idea is that, to track a object through frames, we can obtain multiple different association results from a model by varying the frames it can observe, i.e., skipping frames in observation.  
As the differences in observations do not alter the identities of objects, the obtained association results should be consistent. 
Based on this rationale, we generate multiple observation paths, each specifying a different set of frames to be skipped, and formulate the Path Consistency Loss that enforces the association results are consistent across different observation paths. We use the proposed loss to train our object matching model with only self-supervision. 
By extensive experiments on three tracking datasets (MOT17, PersonPath22, KITTI), we demonstrate that our method outperforms existing unsupervised methods with consistent margins on various evaluation metrics, and even achieves performance close to supervised methods.\footnote{Code available at \href{https://github.com/amazon-science/path-consistency}{github.com/amazon-science/path-consistency}.}

\end{abstract}

\vspace{-5mm}
\section{Introduction}

Multi-Object Tracking (MOT) is the task of identifying and tracking all object instances present in a video. It benefits a wide spectrum of applications, such as monitoring vehicle traffic, wildlife activity, pedestrians, etc. 
 While there have been several advanced supervised methods developed for  MOT~\cite{motr,MeMOT,fairMOT,centerTrack,trackformer,TuSimple,PermaTr,MotionTrack,FineTrack}, the overall advancement in MOT is slower compared to other vision tasks, mostly due to the prohibitively expensive cost of acquiring high-quality object identities annotations for MOT training. As of today, MOT17~\cite{MOT17}, a dataset of 14 videos, remains one of the most widely used datasets in MOT. 

\begin{figure}[!t]
    \centering
    \includegraphics[width=\columnwidth]{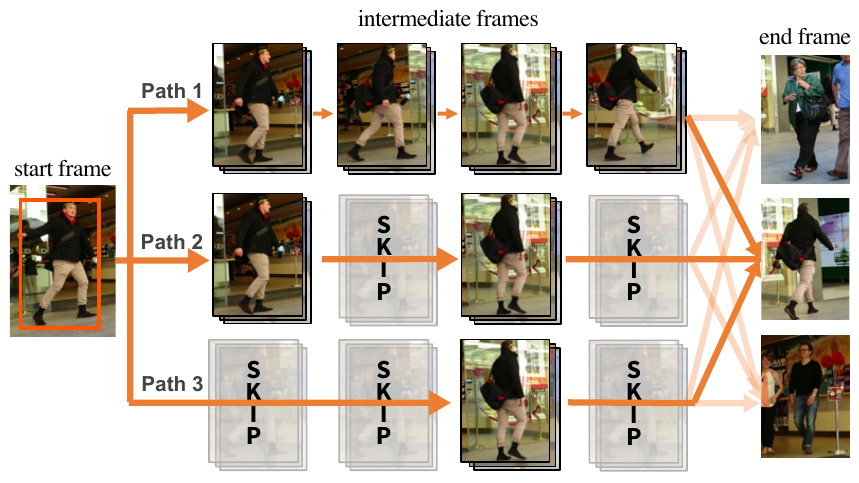}
    \caption{\small 
    We propose a novel concept of path consistency for self-supervised MOT. We define a observation path for an object as a temporal list of observed frames from the start to end frame. As such, for the same object, we can generate multiple paths by skipping intermediate frames. As different observations of the same object does not alter its identity, the association results should be consistent across different paths.
    }
    \label{fig:teaser}
\end{figure}

To overcome this problem, some recent works~\cite{UNS, U2MOT, SimpleReID,UTrack,USCL,UEANET} started shifting their interest on unsupervised learning for MOT. These approaches leverage off-the-shelf object detectors to localize all the object instances in each frame of a video and then learn the temporal object association across frames in an unsupervised manner. 
Multiple works~\cite{SimpleReID,UEANET,U2MOT} propose to learn the object associations by creating pseudo-tracklets to estimate object IDs. 
However, they are vulnerable to the noises in tracklets, especially when the groundtruth tracklet of an object is incorrectly broken into several pseudo-tracklets, each being given a different object ID from the others.
To avoid relying on pseudo-labels, other works~\cite{UTrack, USCL} explore self-supervision with temporal consistency objective, by tracking objects from the current frame to a future frame then back to the current frame while requiring each object is associated back to itself. 
However, these works only consider future frames in very short temporal distances (e.g., 1-2 frames)
where objects lack substantial appearance difference, thus limiting the quality of their object embeddings. 
Meanwhile, 
UNS~\cite{UNS} leverages a motion model and an appearance model to create cross-supervision between their tracking outputs, yet it also only considers matching objects among adjacent frames. 

In fact, the ability of robust object matching over long temporal distances is critical for MOT models as objects often encounter occlusion during tracking, where their bounding boxes are lost for a potentially long period of time, thus cannot be tracked by mere frame-by-frame matching. It requires long-distance matching to re-associate their bounding boxes after occlusion. 
However, very few works have managed to propose reliable solutions that tackle occlusion without using groundtruth supervision. 

In this paper, we introduce the novel concept of path consistency, which can be used as a reliable self-supervised signal to learn a robust object matching model. To elaborate, when tracking an object, we define an observation path of it as a temporal list of observed frames between a start frame and a end frame. As shown in Figure \ref{fig:teaser}, we are able to generate multiple different observation paths for the same object by randomly skipping intermediate frames.  
Based on an observation path, we can compute the temporal association between the object and all other objects from the start frame to the end frame.
As such, path consistency dictates that the temporal associations should be consistent across paths, 
since the identities of objects do not change in different paths.
Based on this rationale, we propose the \textit{Path Consistency Loss} (PCL) that estimates the associations of objects along different paths and trains the object matching model by minimizing the discrepancy among association probability derived from different paths.

Distinct from prior self-supervised methods~\cite{UTrack,UNS,USCL}, our formulation learns object matching over both \textit{short and long distances}. As paths contain frame skipping of varied lengths, the model trained with PCL is able to learn not only short-distance matching (from paths with minimal frame skipping), but also long-distance matching (from paths with consecutive frame skipping) which is essential to handle occlusion during inference. 
In addition, we sample challenging training examples where paths span long temporal horizons, hence enabling our path consistency formulation to learn robust object matching models and handle occlusion.

We evaluate our method on three benchmark tracking datasets, MOT17~\cite{MOT17}, PersonPath22~\cite{personpath22} and KITTI~\cite{KITTI}, and show it successfully achieves new state-of-the-art results over prior unsupervised methods and performs even close to recent supervised methods. We also provide extensive ablation studies to validate the design of our framework and improvement in handling occlusion. 
Our contributions are summarized as follows:
\begin{itemize}
    \item We propose the novel concept of path consistency that can be used as a reliable self-supervised signal to train a robust object matching model. Based on the concept, we formulate the Path Consistency Loss (PCL) that is capable of learning robust long-distance matching, which is essential in handling occlusion and large appearance changes in tracking.
    \item We establish new state-of-the-art performance over prior unsupervised methods on three challenging tracking benchmarks. Extensive ablation studies also validate our improvement in long-distance matching and tracking over occlusion.
    
\end{itemize}

%
%
%

%
%


\section{Related Work}

\myparagraph{Supervised multi-object tracking.}
Most supervised MOT methods~\cite{globalTrackingTransformer, MeMOT, trackformer, fairMOT, centerTrack, siammot, zhang2008global,berclaz2011multiple,zamir2012gmcp,kim2015multiple,tang2017multiple,henschel2017improvements,ristani2018features,sheng2018heterogeneous,tang2017multiple, Liu_2022_CVPR, Liu_2022_CVPR_2,Mayer_2022_CVPR,Ma_2022_CVPR,Yu_2022_CVPR,Li_2022_CVPR,Sun_2022_CVPR} follow the \textit{tracking-by-detection paradigm} where all object instances are first detected in each frame, then they are linked based on their similarities across frames to form tracklets. 
This paradigm benefits from the advances in object detection~\cite{ren2015faster,he2017mask,cai2018cascade,tian2019fcos,zhou2019objects,zhou2021probabilistic} and has two threads of works: offline and online MOT. Offline MOT methods~~\cite{zhang2008global,berclaz2011multiple,zamir2012gmcp,kim2015multiple,tang2017multiple,henschel2017improvements,ristani2018features,sheng2018heterogeneous} focus on finding global optimal associations of objects and are computationally heavy in general. 
Online MOT methods~~\cite{tracktor,globalTrackingTransformer, MeMOT, trackformer, fairMOT, centerTrack, siammot, jde, quasidense, bytetrack, motr} focus on optimizing local optimum on associating detected objects and have real-time speed, e.g., Tracktor~\cite{tracktor} uses bounding box regression to estimate new positions of objects.
Meanwhile, the joint learning of the detection and object matching has made rapid progress in performance~\cite{jde, fairMOT}, e.g.,~\cite{trackformer, motr, MeMOT} introduce transformers and object tokens to associate objects with existing tracklets or to identify new ones. 
Despite the efficacy of supervised methods, they require costly human-annotated labels~~\cite{personpath22, MOT17, MOT20,KITTI}. 
In this work, we learn an online object matching model without labels, which we believe is an important step towards mitigating the limited size of tracking datasets.

\myparagraph{Unsupervised multi-object tracking.} 
\label{sec:related-work-unsup}
This task aims to alleviate the high annotation cost of tracking labels. 
While existing motion trackers~\cite{SORT,bytetrack,OCSORT} do not require labels for training, they cannot leverage the appearance information of objects in tracking.
thus often lose objects in the event of movement change or long occlusion. To leverage appearance information, recent unsupervised methods focused on developing new object matching models~\cite{UNS,USCL,UTrack,U2MOT}. \cite{li2023ovtrack} proposes a setting of open-vocabulary MOT.

Among recent works, pseudo-label methods~\cite{SimpleReID,U2MOT} generate pseudo-tracklets to train their object matching models. SimepleReID~\cite{SimpleReID} and UEANet~\cite{UEANET} obtain pseudo-tracklets from motion trackers without detecting tracklet errors, thus those errors propagate to their models. 
U2MOT~\cite{U2MOT} improves the consistency of object IDs within tracklets, yet is still vulnerable to errors in the early stage of training and, more importantly, errors between tracklets. Specifically, the groundtruth tracklet of an object is often broken into several pseudo-tracklets if it experiences occlusions, where each pseudo-tracklet is then given a different object ID, treated as negative samples of each other.
On the other hand, self-supervised methods~\cite{UNS,UTrack,USCL} derive new unsupervised training objective to replace pseudo-labels. OUTrack~\cite{UTrack} and UCSL~\cite{USCL} utilize temporal consistency, i.e., an object tracked from current frame to a future frame then back to current frame should associate back to itself. However, this objective does not hold if objects disappear among the frames, which limits those methods to only allow future frames at near distances (1-2 frames) to ensure objects likely remain present. Unfortunately, objects lack substantial appearance changes in such short distances, causing their failure to handle occlusion by matching objects over long temporal distances.
We show that object matching learned with short distances cannot generalize to long distances (Table \ref{table:ab-asso-accuracy}). 
UNS~\cite{UNS} creates cross-supervision between motion and appearance trackers yet also learn short-distance matching. 
It discusses occlusion-based supervision yet has failed and discarded it, likely due to its focus on short-distance matching.
Meanwhile, PointId~~\cite{idfree} learns a person embedding model by data augmentation, but it requires pretraining on ReID datasets with groundtruth labels.

Unlike existing methods, we propose a novel self-supervised path consistency loss that simutaneously learns object matching over short to long distances without requiring pseudo or groundtruth labels. Thus we are able to consistently track objects in the event of long occlusion.

\section{Methodology}

\begin{figure*}[!t]
  \centering
  \includegraphics[width=\textwidth]{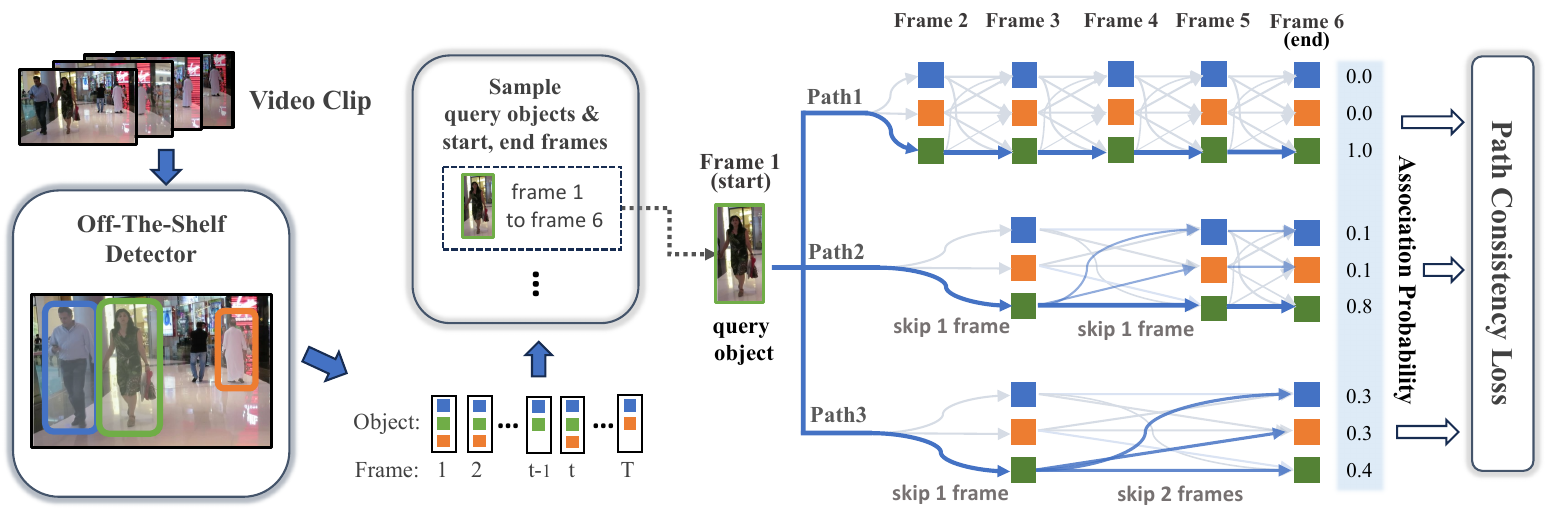}
  \caption{\small \textbf{Overview of Path Consistency Loss (PCL).}
  Our method takes a video clip as input, where objects are localized by an off-the-shelf detector, and uses a selection strategy to choose suitable query objects and their correspondent end frames, then computes PCL to learn association between query objects and objects in end frames. Association probabilities obtained from different paths provide cross-supervision among them and enables self-supervised model learning.
  }
  \label{fig:PCL}
  \vspace{-4mm}
\end{figure*}

\subsection{Framework Overview}
We tackle the unsupervised MOT task with the concept of Path Consistency, which trains a robust object matching model capable of associating objects over long temporal distances, enabling consistent tracking in the event of occlusion.
Following existing unsupervised works~\cite{UNS, tang2017multiple, sheng2018heterogeneous}, we assume the bounding boxes of objects in each video frame are provided by an off-the-shelf detector. Our object matching model takes the detected objects and estimate object embeddings, which encode the appearance and spatial location of each object.
Specifically, let $o^t_i$ denote the $i$-th object in frame $t$ and $h^t_i$ its embedding.
We define the probability that $o^t_i$ matches to an object $o^r_j$ in another frame $r$ as
\begin{align}
 p(o_{i}^{t} \shortrightarrow o_j^{r}) = \frac{\exp(h^t_i \cdot h^r_j)}{\sum_{u=1}^{N_{r}}\exp(h^t_i \cdot h^r_u)},
 \label{eq:matching-probability}
\end{align}
where $N_r$ is the total number of objects in frame $r$. 
Similarly, we also compute $p(o^{r}_{j} \shortrightarrow o_i^{t})$, the matching probability from $o^r_j$ to $o^t_i$.
Importantly, to deal with occlusions and object exit in object association, we add a special $null$ object, $\phi$, to the list of localized objects in each frame. If a real object $o^t_i$ is no longer visible in the following frame $r$, it gets matched to $\phi$.

During inference, our method uses the learned probability $p$ to associate objects to existing tracklets. Given a tracklet $\tau=\{o^{t_m}_m\}^{M}_{m=1}$, consisting of the latest $M$ objects, it estimates the similarity between $\tau$ and a new object $o^t_i$ as 
\begin{equation}
\frac{1}{M} \sum_m \operatorname{gm}\left[ p(o^{t_m}_m \shortrightarrow o^t_i),  p(o^{t}_i \shortrightarrow o^{t_m}_m) \right],
\label{eq:geomean}
\end{equation}
in which, $\operatorname{gm}$ denotes the geometric mean operator. Similar to all other works~~\cite{U2MOT,USCL,UTrack,bytetrack,fairMOT}, we perform bipartite matching based on the tracklet-object similarities to either extend / terminate existing tracklets, or initiate new ones. 

In the absence of tracking annotations, we propose a novel path consistency loss to learn the object matching model over \textit{varied temporal distances with only self-supervision} (Section \ref{sec:PCL}) and two regularization losses to improve ß convergence (Section \ref{sec:regularization}).
As our method is agnostic to model structure and inference procedure, we provide their details in the supplementary materials.

\subsection{Self Supervision with Path Consistency}
\label{sec:PCL}

In this section, we elaborate the concept of path consistency as a reliable self-supervision signal to train a robust object matching model. 
Technically, suppose that we have a query object $o_i^{t_s}$ in the start frame $t_s$, we aim to associating it to its correspondent object, $o_j^{t_e}$, in the end frame $t_e$ ($t_e > t_s$).
Note that the corresponding object $o_j^{t_e}$ is the $null$ object in the case that the query object is not visible in frame $t_e$ (due to being occluded or exiting the scene).

We define an \emph{observation path} as the list of frames a model can observe to compute the association probability distribution between $o_i^{t_s}$ and all visible objects in frame $t_e$. We are able to construct multiple paths by randomly skipping frames, thus creating a different visual observation for each path.  Take the example in Figure \ref{fig:PCL} (right) as illustration, all frames are observed in \mapping{Path1} whereas only a subset of frames are observed in \mapping{Path2} and \mapping{Path3}.
The core principle of the path consistency is, despite the different observations in the paths, the identities of objects remain unchanged, thus the association probability between $o_i^{t_s}$ and $o_j^{t_e}$ should always be higher than all other alternatives. That is, the association probability distribution between $o_i^{t_s}$ and all visible objects in frame $t_e$ should be consistent across paths. Based on this principle, we introduce our path consistency loss in Section \ref{sec:pcl formulation}, and then present our method to sample frames $(t_s, t_e)$ in Section \ref{sec:sampling strategy}.

\vspace{-2mm}

\subsubsection{Path Consistency Loss}
\label{sec:pcl formulation}
Path Consistency Loss (PCL) computes the association probability distributions from all possible paths, $\{ q_\pi | \pi \sim \Pi \}$, and enforces consistency among them\footnote{When the number of paths is large, we conduct Monte Carlo estimation by computing the loss with a smaller set of sampled paths.}, where $\Pi$ is the set of all paths and $q_\pi$ is the association probability distribution obtained using a specific path $\pi=\{t_s, t_1, ..., t_K, t_e\}$. $t_k$ is an intermediate frame in the path. 
We use $q_\pi^{t_k}(j)$ to denote the probability that query object $o^{t_s}_i$ matches with the $j$-th object at frame $t_k$.\\

\myparagraph{Computing Association.} 
To obtain association probability $q_\pi$, we aggregate the matching probability $p$ defined in Eq.\ref{eq:matching-probability} according to the path $\pi$. We have
{\small
\begin{align}
    q_\pi^{t_k}(j) = 
    \begin{cases}
    p(o^{t_s}_i \shortrightarrow o^{t_1}_j), & \text{if } t_k = t_1 \\[5pt]
    \sum_u ~ q^{t_{k-1}}_\pi(u) \cdot p(o^{t_{k-1}}_u \shortrightarrow o^{t_k}_j),            & \text{otherwise } 
    \end{cases}
\end{align}
}
\hspace{-1mm}where the first case defines the association from query object to objects in the first intermediate frame $t_1$, which is the matching probability, $p$, between the objects.
The second case extends the association from a frame $t_{k-1}$ to $t_{k}$ by jointly considering the matching probability from \emph{all objects} in $t_{k-1}$ to $o^{t_k}_j$. In contrast, pseudo-label methods~\cite{SimpleReID,UEANET,U2MOT} only consider the most likely matching, leading to error accumulation.
Overall, we have the association probability between query object and objects in end frame as 
\begin{align}
    q^{t_e}_\pi(j) = \sum_u p(o^{t_s}_i \shortrightarrow o^{t_1}_u) \cdots \sum_v p(o^{t_K}_v \shortrightarrow o^{t_e}_j).
\end{align}
which cumulatively multiples the matching probabilities according to the frames in path.

To reduce the space of possible associations and facilitate learning, we constrain that only two spatially close objects between two neighboring frames can be matched.
We define a binary spatial constraint mask $C^{t_{k-1}, u}_{t_k,v}$, which equals to zero if $o^{t_k}_v$ is one of the $S$ furthest objects to $o^{t_{k-1}}_u$.
Thus, we update $q^{t_e}_\pi$ as 
\begin{align}
   \frac{1}{Z} \sum_u (C^{t_s, i}_{t_1,u}) p(o^{t_s}_i \shortrightarrow o^{t_1}_u) \cdots  \sum_v (C^{t_K, v}_{t_e,j}) p(o^{t_K}_v \shortrightarrow o^{t_e}_j),
\label{eq:new-pcl}
\end{align}
where $Z$ is a normalization factor to ensure the probability $q^{t_e}_\pi$ sums to 1. \\

\myparagraph{Computing Loss.}
After computing the association probabilities for all paths $\{q_\pi^{t_e} | \pi \sim \Pi \}$,  
We define the path consistency loss to enforce consistency among them to learn our model with self-supervision,
\begin{align}
    \mathcal{L}_\text{PC}\left( o^{t_s}_i, t_e \right)
    =\frac{1}{|\Pi|} \sum_\pi KL(q_\pi^{t_e} || \hat{q}) + H(q_\pi^{t_e}),
\label{eq:pcl}
\end{align}
which includes two terms.
The first term is the KL Divergence between each association probability $q_\pi^{t_e}$ and the average of all probabilities, $\hat{q} = \sum_\pi q_\pi^{t_e} / |\Pi|$.
Thus, it is minimized only when \textit{all probabilities follow the same distribution}.
The second term minimizes the entropy, $H(\cdot)$, of each probability to require only one object in frame $t_e$ can be associated to the query object, thus avoids the trivial solution where all $q_\pi^{t_e}$ are consistent yet being uniform distributions.

Our path consistency loss has the key advantages of \textbf{\textit{(i)}} linear complexity w.r.t the number of paths as we compare each $q_\pi^{t_e}$ to the averaged probability $\hat{q}$, hence avoid computing pairwise discrepancy between all $q_\pi^{t_e}$.
\textbf{\textit{(ii)}} $\hat{q}$ reflects association consistency and modulates the loss. 
When the consistency is low, multiple objects can have high probabilities in $\hat{q}$, indicating different association results. 
Unlike pseudo-label methods~\cite{SimpleReID,UEANET,U2MOT} that will enforce a specific association, PCL allows exploring all potential associations by using $\hat{q}$ in $KL(\cdot)$, while $\hat{q}$ also gives higher weights to the more likely ones.

\subsubsection{Selecting Frame Pairs to compute PCL }
\label{sec:sampling strategy}

In principle, based on our PCL formulation we can sample any two random frames ($t_s, t_e$) from a video clip, and then use all objects in $t_s$ as  query objects. However, we observe such unconstrained frame sampling leads to most query objects not visible in the end frames. The unbalance between visible and disappeared query objects causes the model to always matching objects to the $null$ objects and makes the training hard to converge. 

To constrain the frame sampling, we pick frame pairs ($t_s, t_e$) under the condition that one object in $t_s$ should remain visible in intermediate and end frames to serve as the query object while $t_s$ and $t_e$ should be temporally disclose to support learning long-distance matching.
As such, we develop a simple heuristics that uses the temporal chain of the IOU similarity to reject most frame pairs. With the existing solution, we observe that the query objects are present in end frames $98\%$ of all sampled pairs. We include the details of our approach in the supplementary details.
Notice that we can also employ more advanced approaches to include more challenging training examples and reject more noisy frame pairs, which we will leave for future works.

\subsection{Regularization Losses}
\label{sec:regularization}
We further introduce the following two losses to regularize the model training together with our path consistency loss.

\myparagraph{One-to-one matching loss.} 
To avoid that the path for two different query object cross each other, we constrain that a real, non-$null$ object in frame $r$, $o_j^r$, can not be matched by more than one real objects in frame $t$. As elaborated earlier, the probability $p(o^t_i \shortrightarrow o^r_j)$ will be close to $1$ between two matched real objects $o_i^t$ and $o_j^r$. As such, we compute $ \sum_i p(o^t_i \shortrightarrow o^r_j)$ to approximate the number of real objects in frame $t$ that are matched to object $o^r_j$ and penalize the value if it exceeds 1. Mathematically, our loss is defined as:
\begin{equation}
    \mathcal{L_{\text{OM}}} = \frac{1}{T^2} \sum_{t, r} \frac{1}{N^r} \sum_{j} \max\left(1, \sum_{i} p(o^t_i\shortrightarrow o^r_j) \right),
\label{eq:match}
\end{equation}
which penalizes any many-to-one real object matching between every two frames $(t, r)$ in the input video clip of length $T$. $N^r$ is the number of real objects in frame $r$.

\myparagraph{Bidirectional consistency loss.}
In our model, we compute matching probability from frame $t$ to a later frame $r$ (forward in time) but also probability from frame $r$ to frame $t$ (backward in time). 
To ensure the matching probabilities are invariant to the temporal directions, 
we design a bidirectional consistency loss as: 
{
\begin{equation}
\mathcal{L_\text{BC}} =\frac{1}{T^2 N_t N_r} \sum_{t,r} \sum_{i,j} || p(o^t_i \shortrightarrow o^r_j) - p(o^r_j \shortrightarrow o^t_i) || ^2.
\label{eq:forward-backward}
\end{equation}
}
\hspace{-1mm}Given Eq. \ref{eq:match} and Eq. \ref{eq:forward-backward}, our overall model objective is the sum of path consistency loss and two regularization losses: $\mathcal{L} = \mathcal{L_{\text{PC}}} + \mathcal{L_{\text{OM}}} + \mathcal{L_\text{BC}}$, 
where we combine all loss terms with a balance weight of 1.0.

\section{Experiments}

\subsection{Experimental setup}

\myparagraph{Datasets.} We conduct experiments on three tracking datasets: \textbf{MOT17}~\cite{MOT17} is a widely adopted person tracking dataset containing 14 video sequences, featuring crowded scenes in indoor shopping malls or outdoor streets. The videos are split into 7 training sequences and 7 test sequences.
\textbf{PersonPath22}~\cite{personpath22} is a large-scale pedestrian dataset with 236 videos, which are collected from varied environment settings that contain \textit{longer occlusions and more crowded scenes}. 
It is split into 138 training videos and 98 test videos.
\textbf{KITTI}~\cite{KITTI} is person and car tracking dataset, featuring videos with lower frame rates. We follow UNS~\cite{UNS} to report performance on car tracking.

\myparagraph{Evaluation metrics.} Following the evaluation protocols of the datasets, we report our results with several metrics: HOTA (DetA, AssA, LocA), MOTA (IDsw, FP and FN) and IDF1. \textit{We emphasize on HOTA, AssA, IDsw and IDF1} as they measure association accuracy. Other metrics are more influenced by the detection accuracy of detectors.

\myparagraph{Implementation details.} 
Our model input is a video clip of $T=48$ frames, which we found balances between performance and efficiency.
For PCL, we sample at most $G=25$ different paths for each query object and end frame and set $S=\sqrt{\bar{N}}$ in spatial mask constraint, where $\bar{N}$ is average number of objects per frame. 
We implement our model with Pytorch and trained with Adam Optimizer on Tesla V100 and a learning rate of 0.0001.  
In inference, we use the latest $M=4$ bounding boxes of each tracklet to estimate its association with new objects and maintain unmatched tracklets for 30 frames.
We provide more implementation details in the supplementary materials.

\subsection{Comparison to the state-of-the-art}

We compare our method with recent unsupervised object matching methods, where SimpleReID~\cite{SimpleReID}, U2MOT~\cite{U2MOT} are \textit{pseudo-label methods} and  OUTrack~\cite{UTrack}, UNS~\cite{UNS}, UCSL~\cite{USCL} are \textit{self-supervised methods}.
For complete comparison, we include the results of motion-based methods~\cite{SORT,bytetrack,OCSORT}, as they do not require object ID annotation, and also results of recent supervised methods.

\begin{table}[!t]
   \centering
   \resizebox{0.95\columnwidth}{!}{%
\begin{tabular}{lccccc}
\toprule
& Sup.  & HOTA $\uparrow$   & IDF1$\uparrow$  & MOTA$\uparrow$  & IDsw$\downarrow$               \\ 

\midrule
\multicolumn{6}{c}{Private Detection}   \\ 
\midrule

TransTrack \cite{transtrack}     & $\checkmark$    & 54.1       & 63.9       & 74.5       & 3663       \\
TrackFormer \cite{trackformer}   & $\checkmark$    & 57.3       & 68.0       & 74.1       & 2829       \\
MeMOT \cite{MeMOT}     & $\checkmark$    & 56.9       & 69.0       & 72.5       & 2724       \\
MOTR \cite{motr}       & $\checkmark$    & 57.8       & 68.6       & 73.4       & 2439       \\
MeMOTR \cite{MeMOTR}   & $\checkmark$    & 58.8       & 71.5       & 72.8       & 1902       \\
FairMOT \cite{fairMOT} & $\checkmark$    & 59.3       & 72.3       & 73.7       & 1074       \\
FineTrack \cite{FineTrack}       & $\checkmark$    & 64.3       & 79.5       & 80.0       & 1270       \\
UTM \cite{UTM}         & $\checkmark$    & 64.0       & 78.7       & 81.8       & 1431       \\
MotionTrack \cite{MotionTrack}   & \checkmark & 65.1       & 80.1       & 81.1       & 1140       \\
SimpleReID \cite{SimpleReID}     & $\times$        & 50.4       & 60.7       & 69.0       & -\\
OUTrack \cite{UTrack}  & $\times$        & -& 70.2       & 73.5       & 4110       \\
PointID \cite{PointID} & $\times$        & -& 72.4       & 74.2       & 2748       \\
ByteTrack \cite{bytetrack} $\dag$& $\times$        & 63.1       & 77.3       & 80.3       & 2196       \\
OC-SORT \cite{OCSORT} $\dag$     & $\times$        & 63.2       & 77.5       & 78.0       & 1950       \\
UCSL \cite{USCL}       & $\times$        & 58.4       & 70.4       & 73.0       & -\\
U2MOT \cite{U2MOT} $\dag$        & $\times$        & 64.2       & 78.2       & 79.9       & \textbf{1506} \\ 

\hdashline
\textbf{\textit{Ours}} $\dag$       & $\times$        & \textbf{65.0} & \textbf{79.6} & \textbf{80.9} & 1749       \\ 

\midrule

\multicolumn{6}{c}{Public Detection}    \\ 
\midrule
Tracktor++ \cite{tracktor} $ \ddag $       & $\checkmark$    & 42.1       & 52.3       & 53.5       & 2072       \\
MHT-BLSTM \cite{BLSTM} & $\checkmark$    & 41.0       & 51.9       & 47.5       & 2069       \\
FAMNet \cite{FAMNet}   & $\checkmark$    & -& 48.7       & 52.0       & 3072       \\
LSST \cite{LSST}       & $\checkmark$    & 47.1       & 62.3       & 54.7       & 1243       \\
GSM \cite{GSM}         & $\checkmark$    & 45.7       & 57.8       & 56.4       & 1485       \\
CenterTrack \cite{centerTrack} & $\checkmark$ & - & 60.5 & 55.7 & 2540 \\
TrackFormer \cite{trackformer} & $\checkmark$ & - & 62.3 & 57.6 & 4018 \\
SORT \cite{SORT} $ \ddag $       & $\times$        & 34.0       & 39.8       & 43.1       & 4852       \\
UNS \cite{UNS} $ \ddag $         & $\times$        & 46.4       & 58.3       & 56.8       & 1320       \\

\hdashline      

\textbf{\textit{Ours}} $ \ddag $ & $\times$        & \textbf{49.0} & \textbf{61.2} & \textbf{58.8} & \textbf{1219} \\ 
\bottomrule
\end{tabular}

   }
\caption{\small {\bf Comparison on MOT17 \cite{MOT17} dataset.} $\uparrow$/$\downarrow$: higher/lower is better. ``\textit{Sup.}'' means requiring supervision. ``$\dag$" denotes using YoloX detection. ``$ \ddag $'' denotes using Tracktor++ detection preprocessing.
   The best results of unsupervised methods are in bold. 
   }
   \label{table:mot17}
\end{table}

\myparagraph{Comparison on MOT17.} In Table \ref{table:mot17}, we report the results on MOT17 dataset with both private and public detections. For fair comparison, we use YoloX detections from~\cite{bytetrack} for private detection and follow the detection pre-processing of UNS~\cite{UNS} for public detection. 
Our method outperforms all unsupervised works on both private and public detections, and even exceeds recent fully-supervised methods, e.g., UTM~\cite{UTM}, FineTrack~\cite{FineTrack}.

Comparing with \textit{pseudo-label methods}, we surpass the best competitor, U2MOT~\cite{U2MOT}, by \textit{0.8\%} in HOTA, \textit{1.4\%} in IDF1 and \textit{1\%} in MOTA with the same YoloX detection, as it is vulnerable to errors in pseudo labels. In contrast, our PCL naturally measures model confidence with $\hat{q}$ and avoids enforcing incorrect, unconfident associations. 
Meanwhile, we also exceed \textit{self-supervised methods}~\cite{UTrack,USCL,UNS} as they only learn short-distance matching among adjacent frames. Our method learns object matching over both short and long distances thus is robust to appearance change and occlusion.
Using the same public detection as UNS~\cite{UNS}, we surpass it by \textit{2.6\%} in HOTA, \textit{2.9\%} in IDF1 and \textit{2\%} in MOTA. 
Some works~\cite{SimpleReID,UTrack} use better detectors to improve public detection, hence their performance are not comparable to us.

{
\setlength{\tabcolsep}{7pt}
\renewcommand{\arraystretch}{1.2}
\begin{table}[!t]
\centering
\resizebox{\columnwidth}{!}{%
\begin{tabular}{lcccccc}
\toprule
 & Sup. &  IDF1$\uparrow$ &MOTA$\uparrow$ & IDsw $\downarrow$ & FP$\downarrow$ & FN$\downarrow$ \\ 
\midrule

CenterTrack \cite{centerTrack} & $\checkmark$  &  46.4 &59.3 & 10319 & 24K & 72K \\ 
SiamMOT \cite{siammot} & $\checkmark$ &  53.7 &67.5 & 8942 & 13K & 63K \\ 
FairMOT \cite{fairMOT} & $\checkmark$ &  61.1 &61.8 & 5095 & 15K & 80K \\ 
TrackFormer \cite{trackformer} & $\checkmark$ &  57.1 &69.7 & 8633 & 23K & 47K \\ 

SORT \cite{SORT} & $\times$ &  39.6 &53.4 & 9927 & 2K & 110K \\ 
UNS \cite{UNS} & $\times$ &  41.8 &55.5 & 13877 & 17K & 86K \\ 
PointID \cite{PointID} & $\times$ &  63.1 &68.6 & 6148 & 9K & 66K \\ 

ByteTrack \cite{bytetrack} * & $\times$ &  57.2 &64.6 & 5530 & 9k & 71k \\ 
ByteTrack \cite{bytetrack} $\dag$ & $\times$ &  66.8 &75.4 & 5931 & 17k & 40k \\ 
OC-SORT \cite{OCSORT} $\dag$ & $\times$ &  68.2 & 74.0 & 4931 & 17k & 41k \\ 
\hdashline
\textit{\textbf{Ours}}* & $\times$ &  60.5 &64.8 & 5008 & 7k & 73k \\
\textit{\textbf{Ours}} $\dag$  & $\times$ &  \textbf{69.4} &\textbf{75.8} & \textbf{4663} & \textbf{15k} & \textbf{38k} \\ 
\bottomrule

\end{tabular}
}

\caption{\small {\bf Comparison on PersonPath22  \cite{personpath22} dataset.} $\uparrow$/$\downarrow$: higher/lower is better. The best results of unsupervised methods are in bold. ``*" denotes using the public detection of Personpath22 provided by the authors. ``$\dag$" denotes using YoloX detections. 
}
\label{table:pp}
\end{table}
}

{
\setlength{\tabcolsep}{7pt}
\renewcommand{\arraystretch}{1.2}
\begin{table}[!t]
\centering
\resizebox{0.95\columnwidth}{!}{%
\begin{tabular}{lccccc}
\toprule
& Sup. & HOTA$\uparrow$  & AssA$\uparrow$ & DetA$\uparrow$ & MOTA$\uparrow$ \\ 
\midrule

FAMNet \cite{FAMNet}  & $\checkmark$ & 52.6 & 45.5 & 61.0 & 75.9 \\
mmMOT \cite{mmMOT}  & $\checkmark$ & 62.1 & 54.0 & 72.3 & 83.2 \\
CenterTrack \cite{centerTrack}  & $\checkmark$ & 73.0 & 71.2 & 75.6 & 88.8 \\
LGM \cite{LGM}  & $\checkmark$ & 73.1 & 72.3 & 74.6 & 87.6 \\
TuSimple \cite{TuSimple}     & $\checkmark$ & 71.6 & 71.1 & 72.6 & 86.3 \\
PermaTr \cite{PermaTr}      & $\checkmark$ & 78.0 & 78.4 & 78.3 & 91.3 \\
SORT \cite{SORT}      & $\times$     & 42.5 & 41.3 & 44.0 & 53.2 \\
UNS \cite{UNS}& $\times$     & 62.5 & 65.3 & 61.1 & -    \\
OC-SORT \cite{OCSORT}& $\times$     & 76.5 & 76.4 & 77.2 & 90.8 \\ 

\hdashline

\textbf{\textit{Ours}} & $\times$     & \textbf{78.8} & \textbf{80.3} & \textbf{77.9} & \textbf{91.0} \\ 
\bottomrule

\end{tabular}
}

\caption{\small {\bf Comparison on KITTI \cite{KITTI} dataset (Tracking Car).} $\uparrow$/$\downarrow$: higher/lower is better. The best results of unsupervised methods are in bold. We use the same detections as OC-SORT. 
}
\label{table:kitti}
\end{table}
}

\myparagraph{Comparison on PersonPath22.} 
In Table \ref{table:pp}, we report results on PersonPath22 dataset, which is more challenging than MOT17 as \textit{its videos contain more crowded scenes with long occlusions}.
We compare with both public and private detections and successfully outperform ByteTrack by \textit{3.3\%} and \textit{2.6\%} in IDF1 respectively, also reducing IDsw by large margins. The larger improvement in IDF1 demonstrates the better ability of our model in matching over long temporal distances and re-associate objects over occlusion to deliver more consistent tracking results. 
With even public detection, our model has achieved better performance than CenterTrack~\cite{centerTrack}, TrackFormer~\cite{trackformer} and surpasses FairMOT~\cite{fairMOT} in MOTA, despite being unsupervised. These results show that our model effectively benefits from more training data, exposing the possibility of large-scale self-supervised learning with enormous tracking videos available online.

More importantly, in Table \ref{tab:breakdown}, we study model performance under challenging scenarios, including different occlusion length (measured by frames), object moving speeds and people density. Our model consistently outperform the best competitor, ByteTrack, by large margins, thanks to our PCL learning more robust object matching to occlusion and appearance changes.

\myparagraph{Comparison on KITTI.} 
In Table \ref{table:kitti}, we further report our performance on KITTI dataset for tracking car, following the evaluation protocol of UNS~\cite{UNS}. KITTI features varied moving velocity of objects and lower frame rate on average, thus objects have large appearance and location changes in adjacent frames~\cite{OCSORT}. 
We use the shared detections from OC-SORT~\cite{OCSORT}.

Our model exceeds the best competitor, OC-SORT~\cite{OCSORT}, by \textit{2.3\%} in HOTA, \textit{3.9\%} in AssA, \textit{0.7\%} in DetA and \textit{0.2\%} in MOTA. 
This is because cars have changes in their moving speeds and bounding box aspect ratios during driving (e.g., when making turn), which leads to low overlap between bounding boxes of the same object in consecutive frames and failure of pure motion trackers~\cite{SORT,OCSORT}. 
On the other hand, our method learns object embeddings to encode object appearance thus successfully links cars in various driving scenarios, further validating our learned object matching is robust to appearance and location changes.

\subsection{Ablation Study}

We conduct ablation study to examine our key design choices. We train our unsupervised models on MOT17 test set and evaluate on MOT17 train set, since the annotations of the former set are not publicly available. Finally, we use the publicly released SDP detections.

\begin{table}

\centering

\resizebox{0.8\columnwidth}{!}{
\begin{tabular}{lccccc} 
\toprule
& \multicolumn{5}{c}{Occlusion Length (frames)} \\
    & 0 & 1-20 & 20-40 & 40-60 & 60+ \\
\cmidrule{2-6} 
 UNS \cite{UNS} & 49.2 & 36.8 & 32.8 & 27.0 & 26.1 \\
 OC-SORT \cite{OCSORT} & 77.9 & 61.9 & 50.1 & 39.3 & 39.2 \\
 \textbf{\textit{Ours}} & \textbf{79.2} & \textbf{64.3} & \textbf{50.7} & \textbf{40.7} & \textbf{39.8}  \\ 
\bottomrule
\end{tabular}
}

\vspace{3mm}

\centering
\resizebox{0.9\columnwidth}{!}{
\begin{tabular}{lccccc} 
\toprule
& \multicolumn{3}{c}{Movement Speed} & \multicolumn{2}{c}{People Density} \\ 
    &  slow & medium & fast & low & high\\ 
\cmidrule{2-4} \cmidrule(lr){5-6}
UNS \cite{UNS} & 42.7 & 40.3 & 38.3 & 44.7 & 42.4 \\
OC-SORT \cite{OCSORT} & 65.9 & 63.6 & 61.7 & 69.7 & 64.8  \\

 \textbf{\textit{Ours}} & \textbf{68.3} & \textbf{65.1} & \textbf{62.6} & \textbf{71.4} & \textbf{66.3}  \\ 
\bottomrule
\end{tabular}
}

\caption{\small \textbf{Model Performance (IDF1) under different challenging scenarios on PersonPath22}. }
\label{tab:breakdown}
\end{table}




{
\setlength{\tabcolsep}{7pt}
\renewcommand{\arraystretch}{1.1}

\begin{table}[!t]
\centering
\resizebox{0.96\columnwidth}{!}{%
\setlength{\tabcolsep}{10pt}
\begin{tabular}{cccccc}
\toprule
$|t-r|$ & 1-4 & 5-8 & 9-16 & 17-32 & 32-48\\ 
\midrule
PCL $(s=T)$ & 99.5 & 99.1 & 98.1 & 96.3 & 94.5 \\
\midrule
s=32 & 99.5 & 99.2 & 98.2 & 96.3 & 93.0 \\ 
s=24 & 99.5 & 99.1 & 98.0 & 94.8 & 88.0 \\
s=16 & 99.6 & 99.1 & 98.0 & 91.4 & 81.7 \\
s=8 & 99.5 & 99.1 & 95.7 & 84.6 & 71.3 \\ 
s=4 & 99.5 & 98.0 & 91.8 & 78.3 & 64.1 \\ 
\bottomrule
\end{tabular}

}
\vspace{-2mm}
\caption{ \small
\textbf{Accuracy of the matching probability} $p(o^t_i \shortrightarrow o^r_j)$. 
Our method has learned accurate object matching over short to long temporal distances.
}
\label{table:ab-asso-accuracy}
\end{table}
}

{
\setlength{\tabcolsep}{7pt}
\renewcommand{\arraystretch}{1.1}
\begin{table}[!t]
\centering
\resizebox{0.95\columnwidth}{!}{%
\begin{tabular}{cccccccc}
\toprule
& L=0& L=10& L=20& L=30& L=40 & L=50 & L=60\\ 
\toprule
UNS & 64.0 & 62.5 & 58.0 & 55.0 & 53.5 & 52.0 & 50.0 \\ 
OC-SORT & 67.7 & 66.9 & 65.8 & 65.1 & 63.9 & 63.5 & 62.5 \\ 
\midrule
\textbf{\textit{Ours}} & \textbf{68.9} & \textbf{68.3} & \textbf{67.4} & \textbf{66.9} & \textbf{66.1} & \textbf{65.9} & \textbf{65.3} \\ 
Gain & {\color{orange} \textbf{+1.2}} & {\color{orange} \textbf{+1.4}} & {\color{orange} \textbf{+1.6}} & {\color{orange} \textbf{+1.9}} & {\color{orange} \textbf{+2.2}} & {\color{orange} \textbf{+2.4}} & {\color{orange} \textbf{+2.8}} \\ 
\bottomrule
\end{tabular}
}
\caption{\small {\bf Comparison of IDF1 for tracking over occlusion.} When increasing the minimal occlusion length (number of frames), our model has more robust performance and exceeds the best competitor OC-SORT by large margins (as shown in last row).}
\label{table:extend-occlusion}
\end{table}
}


\myparagraph{Has the model learned long-distance object matching?}
We exam the accuracy of our learned matching probability, $p(o^t_i \shortrightarrow o^r_j)$, where $p$ is correct if $o^r_t$ is the correct matching of $o^t_i$ and has the largest matching probability. 
In the \textit{first two rows} of Table \ref{table:ab-asso-accuracy}, we show the accuracy and break it down by the temporal distances between objects $|t-r|$. 
It shows our matching probability is robust to temporal distances. As $|t-r|$ increases, the accuracy drops only by small margins and remains \textit{94.5\%} even at the distance of 32-48 frames,
validating our model has learned object matching from short to long distances.

\begin{figure*}[!t]
    \centering
    \includegraphics[width=0.9\textwidth]{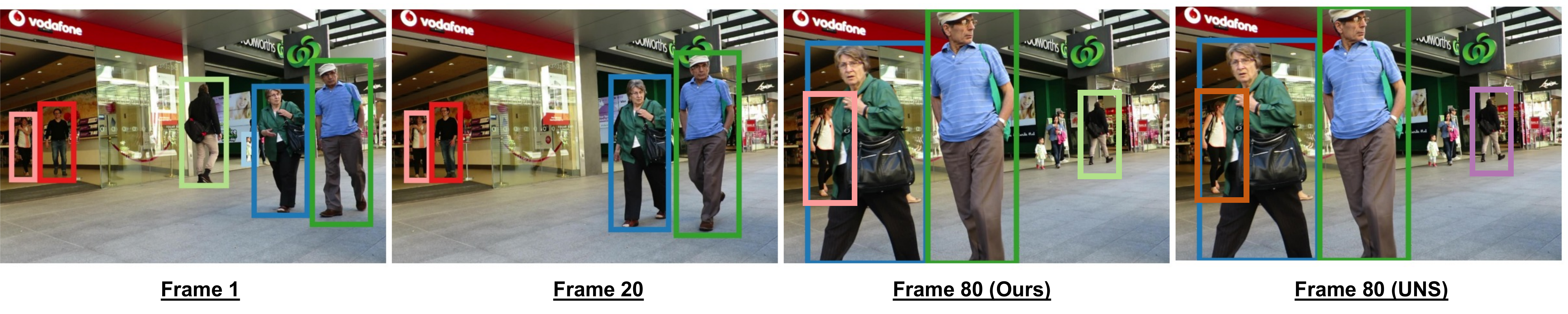}
    \vspace{-3mm}
    \caption{\textbf{Qualitative comparison between our model and UNS~\cite{UNS}}.
    We visualize the tracking on three frames. UNS cannot track the person in green bounding once he is occluded and assigns him a new ID (purple) on frame 80. It also fails to track the person in pink bounding box when she is only partially visible in frame 80.
    We can consistently track with both people with the same IDs (green, pink). 
    }
    \label{fig:quantative}
    \vspace{-5mm}
\end{figure*}

\myparagraph{Can the model only learn short-distance matching?} Next, we further exam if learning long-distance matching is necessary and if a model only learned short-distance matching can generalize to long-distance matching?
In rows 2-6 of Table \ref{table:ab-asso-accuracy}, we limit the maximal number of continuous frames that can be skipped in PCL, denoted by $s$.  
A smaller $s$ means the model is only trained to match objects over short temporal distances.

As $s$ decreases from 32 to 4, while accuracy for short distances (1-8) remains high, it deteriorates quickly for longer distances. Comparing to the first row where we do not limit $s$, accuracy drops by \textit{30.4\%} on the distance of 32-48. It shows that \textit{model trained for short-distance matching cannot generalize to longer distances}, which is the drawback of previous self-supervised methods~\cite{UTrack,UNS,USCL}. Our PCL has successfully learned matching probability over varied temporal distances while not requiring groundtruth annotations. We also observe that 50\% of frame skipping in paths are longer than 8 frames, which boosts our model robustness for long distance matching and the appearance changes caused by it.

\myparagraph{Can the model handle long occlusion?} After validating the accuracy of probability matching, we exam our model ability for handling occlusion at inference.
We notice that most occlusions in MOT17 videos are of short lengths, e.g., less than 5 frames, which cannot reflect a model's ability to handle long-distance matching. Hence, we increase the occlusion length in a video by keep dropping the bounding boxes of objects after occlusion, until the occlusion lengths reach $L$ frames. If the original length of an occlusion is longer than $L$, we do not increase it. Thus, $L=0$ means no occlusion is extended while $L=60$ means all occlusions last for at least 60 frames.

In Table \ref{table:extend-occlusion}, we report IDF1 with different occlusion length $L$ and compare with UNS and one of the state-of-the-art trackers, OC-SORT, as a strong baseline. Since other related methods~\cite{SimpleReID,UTrack,USCL,U2MOT} do not release their codes or models, we are unable to compare with them.
As can be seen, our model consistently provides higher IDF1 for all ranges of occlusion lengths. More importantly, when increasing occlusion lengths, the performance of UNS and OC-SORT decreases quickly, while {our performance is more robust and decreases slowly}, which echos our results in Table \ref{tab:breakdown}.
Our IDF1 at $L=50$ is still higher than the IDF1 of OC-SORT at $L=20$, showing our model can learn to match the same objects over varied temporal distances and is more robust to occlusion. 

{
\setlength{\tabcolsep}{7pt}
\renewcommand{\arraystretch}{1.1}
\begin{table}[!t]
\centering
\resizebox{\columnwidth}{!}{%
\centering
\begin{tabular}{l @{\hskip 1mm}|c|cc|cccc}
\toprule
&$\mathcal{L}_\text{PC}$ & $\mathcal{L}_\text{OM}$ & $\mathcal{L}_\text{BC}$   & IDF1  &HOTA&MOTA  &IDsw\\ 
\toprule

 1 &$\times$ & \checkmark & \checkmark   & 3.4 & 8.3 & 1.4 & 70k \\
 2 & (a) & \checkmark & \checkmark  &  26.9 & 26.8 & 37.8 & 29k \\
 3 & (b) & \checkmark & \checkmark  &  66.6 &59.4 & 63.9 & 329 \\ 
\midrule
4 & \checkmark &  $\times$ & \checkmark &  62.7  &57.1 & 63.4  & 591 \\ 
5 & \checkmark & \checkmark & $\times$  &  66.3  &59.8 & 63.6  & 466 \\
\midrule
6 & \checkmark & \checkmark & \checkmark  &  \textbf{68.9}  &\textbf{60.9} &\textbf{63.7}  & \textbf{257} \\ 
\bottomrule

\end{tabular}
}
\caption{\small
\textbf{Ablation study on proposed losses}.
In row 1-3, we demonstrate the effect of our path consistency loss. 
In row 4-5, we show the effect of our regularization losses.
``(a)" denotes randomly sampling start and end frames for paths and ``(b)'' denotes disabling spatial constraint mask in Eq \ref{eq:new-pcl}. 
} 

\vspace{-7mm}
\label{table:ab-loss}
\end{table}
}

\myparagraph{What is the effect of path consistency loss?}
In Table \ref{table:ab-loss} (row 1-3), we exam the results of removing PCL and also two modifications of it: 
(a) randomly sampling the starting and end frames for paths; (b) removing the spatial constraint mask in Eq.\ref{eq:new-pcl}; 
Firstly, without PCL (row 1), the model cannot learn meaningful object matching using only the regularization losses thus the tracking accuracy is close to zeros.
When randomly sampling the starting and end frames (row 2), it causes most query objects are not visible in the end frames, causing imbalance between visible and disappeared objects. Thus, model matches most objects to $null$ objects, also leading to low IDF1.
Next, when removing the spatial constraint mask (row 3), the model may incorrectly match two spatial disclose objects, decreasing IDF1 by 2.3\%. 

\myparagraph{What is the effect of regularization losses?} 
In Table \ref{table:ab-loss} (row 4-5), we ablate the effect of the regularization losses in Section \ref{sec:regularization}. As can be seen, while PCL contributes most to the model performance, the two regularization losses $\mathcal{L}_\text{OM}$, $\mathcal{L}_\text{BC}$ are also helpful for achieving better performance. Since $\mathcal{L}_\text{OM}$ prevents multiple objects matching to the same objects and $\mathcal{L}_\text{BC}$ ensures the consistency between forward and backward matching, removing either of these loss terms leads to around 6.2\% and 2.6\% drop in IDF1, respectively. These results show the collective benefits of using different losses to optimize our unsupervised object tracker. 

\subsection{Qualitative results}

In Figure \ref{fig:quantative}, we visualize the tracking results for one video from MOT17 using our model and UNS~\cite{UNS}, where the color indicates their IDs. Both our model and UNS can track two non-occluded front people from frame 1 to frame 80. However, UNS lose the man in green bounding box after he is occluded and assign a new ID (purple) to him on frame 80. It also cannot associate the woman in the pink bounding box when she is only partially visible in frame 80. Our model correctly associate two people with the same IDs (green and pink). 
This qualitative comparison suggests that our model can better track the object over occlusion.

\section{Conclusion}

To tackle unsupervised MOT task, we proposed a novel concept of path consistency.
We defined a path to track an object as a temporal list of observed frames from the start to the end frame. 
As such, to track the same object, we can generate multiple paths by skipping intermediate frames while the association results should be consistent across different paths. 
We leveraged this consistency in our self-supervised Path Consistency Loss to learn object matching model without groundtruth ID supervision. 
Our PCL especially learns robust long distance matching to tackle occlusion at inference time.
With experiments on three popular datasets, our approach has exceeded existing unsupervised methods and performed on par with recent state-of-the-art supervised methods. 
We provided a comprehensive ablation study to validate the model has learned long-distance matching and is robust under challenging scenarios.

\newpage

{\small
\bibliographystyle{ieeenat_fullname}
\bibliography{egbib}
}


\newpage

\section{Supplementary}

In the supplementary material, we include the details of our model structure (Section \ref{sec:model}), simplification of PCL for easy computation (Section \ref{sec:simplify}), implementation details (Section \ref{sec:imp}), additional ablation study (Section \ref{sec:ab}) and our approach for selecting start, end frames for PCL (Section \ref{sec:sampling}).

\subsection{Model Structure}
\label{sec:model}
As introduced in Section 3.1 of the paper, our model generates object embeddings, $h$, to compute the object matching probability, $p$. In this section, we present the details of our model structure.

\myparagraph{Model Input.} 
Recall that the input to our model at training and inference is a clip of $T$ continuous frames, along with objects detected by an off-the-shell detector. 
Take frame $t$ as an example. It contains $N_t-1$ detected objects and one special $null$ object, $\phi$. 
For the real, detected objects, let $\B^t \in \mathbb{R} ^ {N_t-1, 5}$ denote the four coordinates (top, right, left, bottom) and confidence scores of their bounding boxes and $\I^t \in \mathbb{R} ^ {N_t-1, H, W, 3}$ denote the cropped image patches inside the bounding boxes. For the $null$ object, we learn a fixed embedding vector, as will be explained below. 
Thus, $\{\B^t, \I^t\}$ are the inputs to our model. 

\myparagraph{Model Structure.} Our model computes the embeddings for all objects. First, it computes features for the detected objects based on their visual and spatial information. 
It encodes the visual information via convolution layers,
\begin{align}
    \V^t \in \mathbb{R} ^ {N_t-1, D} = \operatorname{convolution}(\I^t),
\end{align}
where $\V^t$ is obtained visual embedding and $D$ is the embedding dimension size. Next, it concatenates $\V^t$ with the spatial information $\B^t$ to obtain the joint embedding, $\F^{t} \in \mathbf{R}^{N_t-1, D+5} = \operatorname{concat}(\V^t, \B^t)$. 
Lastly, we also include a learned embedding of the null object, $f_{\phi}$, to obtain the feature matrix of all objects,  
\begin{align}
    \tilde{\F}^t \in \mathbb{R} ^ {N_t, D+5} = \operatorname{concat}(\F^t, f_{\phi}).
\end{align}

As $\tilde{\F}^t$ is computed for each object individually, 
we further refine the embeddings by considering the context of other objects in the same frame via self-attention layers,
\begin{align}
    \bH^{t} \in \mathbb{R}^{N_t, D} = \operatorname{self-attention}(\tilde{\F}^t),
\end{align}
where $\bH^t$ is the final embeddings of all objects in frame $t$. Then $h^t_i$ (the $i$-th row of $\bH^t$) is the embedding for the object $o^t_i$ in the frame and used to compute matching probability $p(o^t_i \shortrightarrow o^r_j)$ in Eq(1) of the paper.
Note that, we only consider matching from real, non-$null$ object to $null$ object, as searching for the matches of $null$ object is ambiguous.

Overall, our object embedding encodes the visual, spatial information of an object and considers the other objects in the same frame. 
While one can also incorporate the objects in adjacent frames with temporal cross-attention, it increases learning difficulty and cannot converge well given the small scale of existing tracking datasets. 
Therefore, we do not include it in our model.

\subsection{Simplified Formula of PCL}
\label{sec:simplify}
In Eq(6) of the paper, we introduce our path consistency loss $\mathcal{L}_\text{PC}(o^{t_s}_i, t_e)$, 
which contains two terms: KL divergence and probability entropy.
Here we show that the loss can be simplified for easy computation, 
\begin{align}
\begin{aligned}[b]
   \mathcal{L}_\text{PC}\left( o^{t_s}_i, t_e \right)
    &=\frac{1}{|\Pi|} \sum_\pi KL(q_\pi^{t_e} || \hat{q}) + H(q_\pi^{t_e}) \\
    & = H(\hat{q}) - \frac{1}{|\Pi|}  \sum H(q_\pi^{t_e}) + \frac{1}{|\Pi|} \sum H(q_\pi^{t_e})  \\
    & = H(\hat{q}),
\end{aligned}
\end{align}
which is simply the entropy of the averaged association probability distribution, $\hat{q}$.

\subsection{Implementation Details}
\label{sec:imp}
\myparagraph{Model.} We set the length of input video clip as $T=48$. The image dimension is $H=W=64$ and feature dimension $D=64$. 
Specifically, we maintain the aspect ratios of the image patches, resize their longest sides to 64 and add padding to the other sides. Our model uses 6 convolution layers (each with a kernel size of 3x3 and spatial stride of 2) and 2 self-attention layers (each with 8 attention heads). 
Our convolution layers have the same structure as the CNN in UNS.

\myparagraph{Training.} In addition to our PCL and regularization losses, we follow the practices in \cite{PointID,U2MOT,USCL} that improve appearance models via detection techniques. Formally, our method creates two different views of a input video clip via data augmentation (random flip, shift, etc.) and requires the matching probability $p$ to be consistency across the views. It improves convergence speed and avoids trivial solutions.

\myparagraph{Inference.} Similar to all prior works\cite{globalTrackingTransformer,PointID,fairMOT,deepsort,centerTrack,U2MOT,USCL}, we found it is beneficial to combine our learned model with a motion tracker (SORT~\cite{SORT}). Thus, we match new objects to tracklets based on the average of our tracklet-object similarity (see Eq(2) of paper) and IoU score from the motion tracker. 
For computation efficiency, we only maintain the latest $M=4$ object instances in each tracklet. A unmatched tracklet is kept in a buffer for 30 frames to handle occlusion.

\subsection{Ablation Study on Input Modality}
\label{sec:ab}

{
\setlength{\tabcolsep}{7pt}
\renewcommand{\arraystretch}{1.1}
\begin{table}[!t]
\centering
\resizebox{0.8\columnwidth}{!}{%
\centering
\begin{tabular}{cc|cccc}
\toprule
Spatial & Visual & IDF1 & HOTA & MOTA & IDsw \\ 
\toprule
\checkmark & & 62.9 & 57.3 & 62.8 & 1329 \\ 
 & \checkmark & 66.8 & 60.2 & 63.7 & 293 \\
\checkmark & \checkmark & \textbf{68.9}  &\textbf{60.9} &\textbf{63.7}  & \textbf{257} \\ 
\bottomrule
\end{tabular}
}
\caption{Effect of Input Modality. 
}
\label{table:ab-input}
\end{table}
}
Our model associate objects with two input modalities: visual and spatial modalities. 
In Table \ref{table:ab-input}, we compare the effect of removing each modality to study if our model can jointly exploit two modalities. 
We show that removing either input modality leads to a clear performance drop, which also causes slower convergence during training. 
In particular, removing visual information leads to a larger IDF1 drop of 6\% (68.9-62.9) as appearance information is vital in tracking over occlusion. In contrast, using two modalities together leads to the best overall performance, yielding an IDF1 of 68.9\% and a HOTA of 60.9\%. These results suggest that instead of replying on one single modality, our model indeed utilizes the complementary information between the two input modalities to achieve better tracking performance.

\subsection{Selecting Frame Pairs for PCL}
\label{sec:sampling}

As explained in section 3.2.2 of the paper, we compute path consistency loss with sampled start, end frames $t_s, t_e$ and query objects $o^{t_s}_i$. It is important for the sampled data to satisfy that (i) the end frame is as far from the start frame as possible to support learning long-distance matching; (ii) the query object is visible in the intermediate and end frames to obtain meaningful association. 

Unfortunately, the visibility/presence of objects in each frame is unknown in unsupervised setting. Thus, we estimate it with bounding box overlaps (IoU) between objects and derive a sample strategy, which selects query objects to determine the start and end frames, and finds all possible groups of query object, start and end frames in the input video clip to make the best utilization of training data. Note that we only pick one query object in each start frame to have the minimal constraint on selecting the end frame.

Specifically, we start with the first object in frame $1$, i.e., use the object $o^1_1$ as query object and frame $1$ as start frame.
$o^1_1$ is assumed to exist in frame 2 if its IoU with the closest object in frame $2$, $o^{2}_u$, is higher than a threshold, $\sigma$. 
Similarly, the query object exists in frame 3 if $o^{2}_u$ has high IoU with $o^3_v$, its closest object in frame $3$.
We repeat this process until no object with high IoU is found and use the last frame as the end frame.
We also mark $\{o^1_1, o^2_u, o^3_v, \ldots \}$ as \mgrey{used}, signaling they do not need to be re-selected as query objects, as they are likely instances of the same object.
Similarly, we select the remaining objects in frame $1$ and subsequent frames as query objects while skipping the \mgrey{used} objects.
Hence, each unique object in the video clip should be selected as query object for approximately one time while start and end frames are temporally disclose.
With our approach, the median of temporal distances between $(t_s, t_e)$ is 36 frames while query objects are present in 98\% of the end frames. As a result, we observe 50\% of the frame skipping in paths is longer than 8 frames, providing hard training samples for learning long distance matching.

Note the sample strategy does not provide pseudo labels to our model. It only chooses the start and end frames while object associations are learned using PCL. 
Moreover, our method only requires the query object is present up-to the end frame and is not affected if the object is still present after the frame. Yet in pseudo-label methods \cite{SimpleReID,UEANET,U2MOT}, such scenario means the object will form a new tracklet after the frame, thus is assigned with different pseudo IDs before and after the end frame.

\end{document}